\documentclass[preprint,12pt]{elsarticle}

\usepackage[hyphens]{url}  % DO NOT CHANGE THIS
\usepackage{amsmath,amssymb,amsfonts}
\usepackage{algorithmic}
\usepackage{graphicx}
\usepackage{textcomp}
\usepackage{hyperref}
\usepackage{color}
\usepackage{soul}
\usepackage{array}
\usepackage{booktabs}
\usepackage{adjustbox}
\usepackage{caption}
\newcolumntype{P}[1]{>{\centering\arraybackslash}p{#1}}

\usepackage{lineno}
\modulolinenumbers[5]

\usepackage{xcolor}

\journal{Expert Systems with Applications}

\begin{document}
\begin{frontmatter}

\title{SEZ-HARN: Self-Explainable Zero-shot Human Activity Recognition Network}


\author[aff_lab1]{Devin Y. De Silva}
\author[aff_lab1]{Sandareka Wickramanayake}
\author[aff_lab1]{Dulani Meedeniya}
\author[aff_lab2]{Sanka Rasnayaka}

\affiliation[aff_lab1]{
            organization={Department of Computer Science and Engineering},%Department and Organization
            addressline={University of Moratuwa}, 
            city={Moratuwa},
            postcode={10400},
            country={Sri Lanka}}

\affiliation[aff_lab2]{
            organization={School of Computing},%Department and Organization
            addressline={University of Singapore}, 
            city={Singapore},
            country={Singapore}}

%% Abstract
\begin{abstract}
Human Activity Recognition (HAR), which uses data from Inertial Measurement Unit (IMU) sensors, has many practical applications in healthcare and assisted living environments. However, its use in real-world scenarios has been limited by the lack of comprehensive IMU-based HAR datasets that cover a wide range of activities and the lack of transparency in existing HAR models. Zero-shot HAR (ZS-HAR) overcomes the data limitations, but current models struggle to explain their decisions, making them less transparent. This paper introduces a novel IMU-based ZS-HAR model called the Self-Explainable Zero-shot Human Activity Recognition Network (SEZ-HARN). It can recognize activities not encountered during training and provide skeleton videos to explain its decision-making process. We evaluate the effectiveness of the proposed SEZ-HARN on four benchmark datasets PAMAP2, DaLiAc, HTD-MHAD and MHealth and compare its performance against three state-of-the-art black-box ZS-HAR models. The experiment results demonstrate that SEZ-HARN produces realistic and understandable explanations while achieving competitive Zero-shot recognition accuracy. SEZ-HARN achieves a Zero-shot prediction accuracy within 3\% of the best-performing black-box model on PAMAP2 while maintaining comparable performance on the other three datasets.

\end{abstract}

%% Keywords
\begin{keyword}
Human Activity Recognition \sep Zero-shot Learning \sep Inertial Measurement Unit Data.


\end{keyword}

\end{frontmatter}



\section{Introduction}
\label{sec:introduction}
%% Labels are used to cross-reference an item using \ref command.

Human Activity Recognition (HAR) plays a vital role in domains such as remote monitoring~\cite{khojasteh2018improving}, fitness tracking~\cite{p_har_survey}, and remote yoga instruction~\cite{rishan2020infinity}. HAR methods typically rely on either video data or Inertial Measurement Unit (IMU) sensor data. With the growing adoption of wearable devices and advances in sensor technology, IMU-based HAR has emerged as a practical alternative to video-based approaches, particularly in healthcare and remote monitoring applications. However, collecting large-scale labeled IMU datasets is time-consuming and costly, and most existing datasets~\cite{d_daliac,d_utd-mhad,d_pamap2} cover a limited set of activities. Consequently, supervised models trained on such datasets generalize poorly to unseen activities~\cite{p_sshar}.

Zero-Shot Learning (ZSL) offers a solution by enabling models to recognize unseen classes through a shared semantic space built using auxiliary data~\cite{122meedeniya2023DL, p_sshar}. Prior work in Zero-Shot HAR (ZS-HAR)\cite{p_har_zsl_wemb, p_har_zsl_mcross} constructs this space using word embeddings. However, such embeddings often fail to capture the fine-grained motion characteristics that are crucial for distinguishing activities in IMU data. Recently, Tong et al.\cite{p_har_zsl_video} leveraged video data as a more informative modality for building semantic representations.

As IMU data-based HAR is often employed in applications that interact with people, such as patient monitoring \cite{capela2015feature} and ambient-assisted living \cite{zdravevski2017improving}, model explainability is essential to foster user trust. Although some supervised HAR models incorporate post-hoc explanation techniques like SHAP~\cite{p_shap}, Grad-CAM~\cite{selvaraju2017grad}, or attention visualization~\cite{zeng2018understanding}, these methods often produce abstract visualizations that are not intuitive for lay users~\cite{kaur2020interpreting}. Besides, none of the existing IMU data-based ZS-HAR models has explored generating explanations for their decisions. Further, in contrast to explanations for supervised models, explanations for zero-shot models should articulate how unseen activities are recognized using knowledge from seen classes.

\begin{figure}[!t]
\centering
\includegraphics[width=0.75\linewidth]{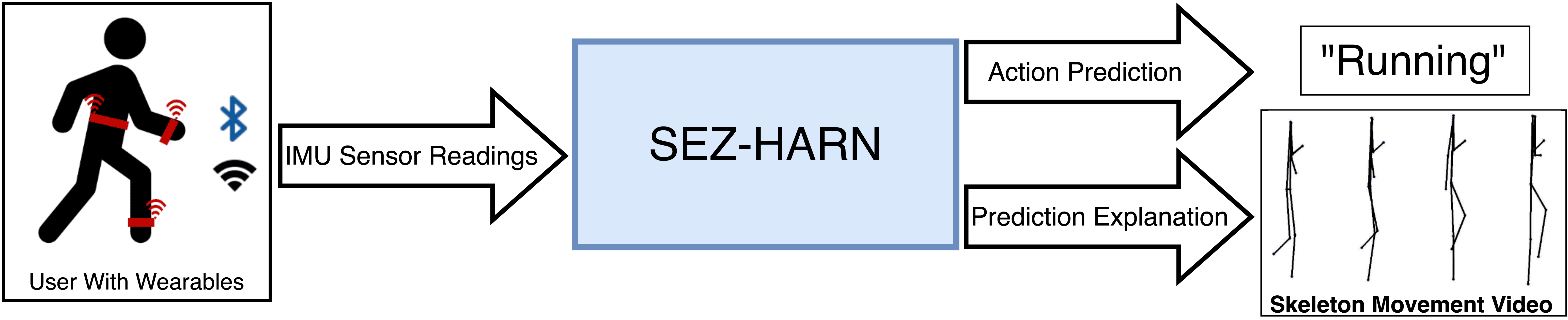}
\caption{Overview SEZ-HARN.} \label{fig:overview_archi}
\end{figure}

\begin{figure*}[!t]
\centering
\includegraphics[width=0.9\textwidth]{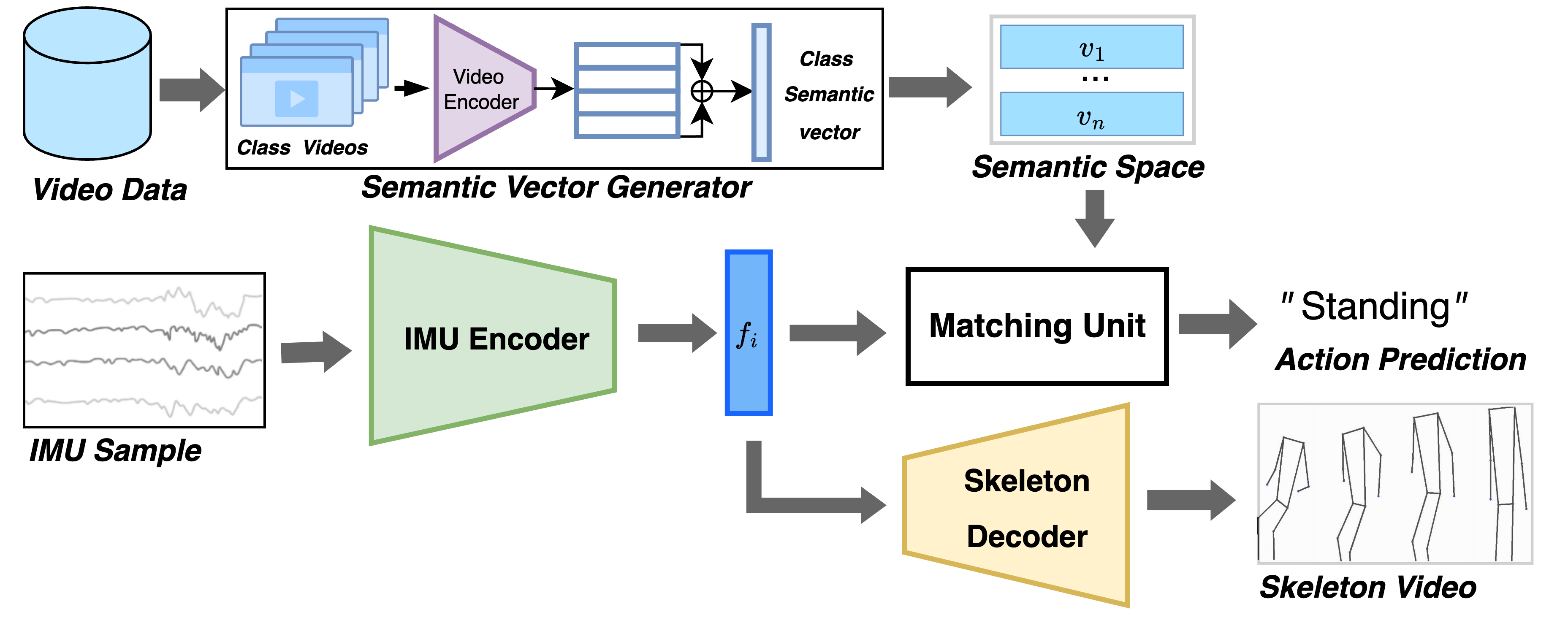}
\caption{The overview of the inference process of SEZ-HARN.} \label{fig:inference_model_diagram}
\end{figure*}

To address these limitations, we propose a novel IMU-based ZS-HAR framework called SEZ-HARN (Self-Explainable Zero-Shot Human Activity Recognition Network). SEZ-HARN leverages auxiliary video data to construct a semantic space enriched with motion information and generates skeleton-based activity videos as intuitive explanations for its predictions [Fig. \ref{fig:overview_archi}]. It comprises a Bi-LSTM encoder to extract temporal patterns from IMU data and a pre-trained video encoder to extract high-level features from auxiliary video data. The output of the video encoder is used to create class semantic vectors representing different activity classes. During training, SEZ-HARN learns to align IMU features with class semantic vectors. During inference, it classifies unseen activities via similarity matching and produces an explanatory skeleton video using a decoder network (see Fig.~\ref{fig:inference_model_diagram}).

We evaluate SEZ-HARN on four public IMU HAR datasets—PAMAP2~\cite{d_pamap2}, DaLiAc~\cite{d_daliac}, UTD-MHAD~\cite{d_utd-mhad}, and MHEALTH~\cite{d_mhealth}. We compare SEZ-HARN with the state-of-the-art black-box ZS-HAR models regarding unseen human activity prediction accuracy and evaluate SEZ-HARN's knowledge transferability from seen to unseen classes. Further, we introduce a new metric for assessing the realism of the generated skeleton movement videos and conduct a user study to assess the human understandability of the generated explanations. Experiment results demonstrate that SEZ-HARN outperforms comparable state-of-the-art black-box ZS-HAR models and generates human-understandable explanations for its decisions.

This paper makes the following contributions.
% \begin{itemize}
%     \item We propose a self-explainable approach for zero-shot human activity recognition using IMU sensory data. Unlike traditional methods that aggregate signals, our approach preserves the original data format and employs Bi-LSTM to effectively handle multivariate time series.
%     \item We present a novel model designed to enhance explainability using sketch-figure movement videos. This approach improves object clarity while substantially reducing the model's size.
%     \item Conducted extensive experiments on the quantitative and qualitative aspects of the proposed method.
%     \item Finally, we propose two new metrics—Dynamic Time Warping (DTW) distance and Discrete Fréchet Distance—to evaluate the understandability and realism of the generated video explanations.
% \end{itemize}

\begin{itemize}
    \item We propose SEZ-HARN, the first IMU-based ZS-HAR framework that integrates explainability by generating skeleton-based activity videos. 
    \item We introduce two new metrics—Dynamic Time Warping (DTW) distance and Discrete Fréchet Distance—to evaluate the understandability and realism of the generated video explanations.
    \item We validate the effectiveness of SEZ-HARN through experiments on four benchmark datasets and a user study assessing the interpretability of its explanations.
\end{itemize}

% The rest of the paper is organized as follows: Section~\ref{sec:relatedWork} reviews existing studies on Human Activity Recognition (HAR) using Inertial Measurement Unit (IMU) sensors, with a focus on Zero-shot learning approaches and model interpretability. Section~\ref{sec:methodology}  describes the the proposed SEZ-HARN model. The experimental study, including data collection, feature engineering, and implementation, are presented in Section~\ref{sec:experiments}. Section~\ref{sec:results} details the results and analysis of our experimental study. The discussion of findings and the future directions are presented in Section~\ref{sec:discussion} Finally, Section~\ref{sec:conclusion} presents concluding remarks.

\section{Related Work}
\label{sec:relatedWork}

\subsection{IMU-based Zero-Shot Human Activity Recognition}
Early research in Zero-Shot Human Activity Recognition (ZS-HAR) relied on expert-defined attribute maps for classification. Cheng et al. \cite{p_har_zsl_alt} introduced an SVM-based approach using binary attribute predictions, later extended by Cheng et al. \cite{p_har_zsl_crf} with a conditional random field and nearest-neighbor classifier. However, these methods were limited by their reliance on manual attribute definitions.

%The focus later shifted toward automated semantic spaces \cite{p_har_zsl_wemb, p_har_zsl_mcross, p_har_zsl_nuactiv}. Matsuki et al. \cite{p_har_zsl_wemb} demonstrated that word embeddings outperformed expert-defined attributes, but they lacked the motion-specific information critical for human activity recognition. To overcome this limitation, Tong et al. \cite{p_har_zsl_video} proposed semantic spaces derived from activity videos, which improved recognition accuracy but failed to capture temporal features or provide explainable predictions. Wu et al. \cite{p_har_zsl_mcross} reframed ZS-HAR as a dual task of classification and latent space regression, offering a novel perspective. Pathirage et al. \cite{125PathirageICONIP2023} advanced this by introducing a Bi-LSTM-based IMU encoding architecture with neighborhood-based unseen class prediction, achieving state-of-the-art performance. However, explainability remains an unresolved challenge across these approaches.

The focus later shifted toward automated semantic spaces \cite{p_har_zsl_wemb, p_har_zsl_mcross, p_har_zsl_nuactiv}. Matsuki et al. \cite{p_har_zsl_wemb} demonstrated that word embeddings outperformed expert-defined attributes. Wu et al.\cite{p_har_zsl_mcross} reframed ZS-HAR as a dual task of classification and latent space regression, offering a novel perspective. Chowdhury et al.\cite{p_zsl_imu_text_projection} utilize textual latent spaces to learn generalized semantics from IMU sensor data using cross-modal contrastive learning, further enhancing performance by integrating sensor context information with motion information. However, textual embeddings lacked the motion-specific information critical for human activity recognition. To overcome this limitation, Tong et al. \cite{p_har_zsl_video} proposed semantic spaces derived from activity videos, which improved recognition accuracy but failed to capture temporal features or provide explainable predictions. Pathirage et al. \cite{125PathirageICONIP2023} advanced this by introducing a Bi-LSTM-based IMU encoding architecture with neighborhood-based unseen class prediction, achieving state-of-the-art performance. However, explainability remains an unresolved challenge across these approaches.

%Table \ref{tab:related_works_comparison} summarizes these methods and their limitations. Matsuki et al. \cite{p_har_zsl_wemb} did not incorporate video attributes or temporal features, while Wu et al. \cite{p_har_zsl_mcross} and Tong et al. \cite{p_har_zsl_video} also failed to model temporal dynamics or provide explanations. Pathirage et al. \cite{125PathirageICONIP2023} combined video attributes with temporal features but lacked mechanisms for explainable decisions.

Our proposed model addresses these limitations by integrating video attributes, temporal features, and self-explainability, offering state-of-the-art performance while ensuring interpretability. This comprehensive approach makes it uniquely suited for safety-critical applications like healthcare monitoring.

\subsection{Explainable Artificial Intelligence}

Explainable AI (XAI) aims to uncover the reasoning behind decisions made by deep learning models, offering transparency and fostering trust. XAI methods can be categorized along several dimensions, one being post-hoc versus ante-hoc approaches. Post-hoc techniques, such as LIME \cite{p_lime} and SHAP \cite{p_shap}, operate externally to trained models, generating explanations after the model has made its predictions. In contrast, ante-hoc methods integrate explainability directly into the model design \cite{wickramanayake2021comprehensible}.

Another categorization differentiates feature attribution explanations and concept-based explanations. Feature attribution methods, such as gradient-based techniques (e.g., SHAP \cite{p_shap} and GradCAM \cite{selvaraju2017grad}), identify influential features driving the model’s decisions. Concept-based explanations, including Concept Activation Vectors \cite{kim2018interpretability} and linguistic explanations \cite{wickramanayake2019flex}, provide higher-level reasoning for model behavior.

Some supervised sensor-based HAR models have adopted XAI to enhance interpretability. For instance, the authors of \cite{p_xai_har} employed post-hoc methods like SHAP, LIME, and Anchors \cite{p_anchors} to explain decisions made by an environmental sensor-based HAR model. Similarly, \cite{p_dexar} converted sensor data into images, and applied existing XAI methods such as Grad-CAM \cite{selvaraju2017grad} and LIME to generate saliency maps highlighting important features at specific time steps. These saliency maps were subsequently translated into text templates to provide explanations comprehensible to non-expert users. However, as both \cite{p_xai_har} and \cite{p_dexar} employ post-hoc methods, their explanations may not fully capture the reasoning behind the model’s decisions.

Self-explainable supervised HAR models offer alternative approaches. For example, the model in \cite{p_dgp} provides explanations by identifying confident and informative sensors, while \cite{zeng2018understanding} uses temporal attention weights to generate heatmaps as visual explanations. However, saliency maps and graphs may still be challenging for lay users to interpret in real-world scenarios \cite{theissler2022explainable,124WickAC2023}. 

Unlike explanations for supervised models, explanations for zero-shot models must go further, elucidating the semantic relationships exploited by the model to recognize unseen classes. This need for semantic clarity highlights a distinct challenge in making zero-shot HAR systems both interpretable and user-friendly.

\section{Methodology}
\label{sec:methodology}

The proposed SEZ-HARN is a ZS-HAR model based on IMU data. It uses video data to establish the semantic relationships between seen and unseen activities and explain its decisions by generating skeleton videos.

\begin{figure*}[!t]
\centering
    \includegraphics[width=\textwidth]{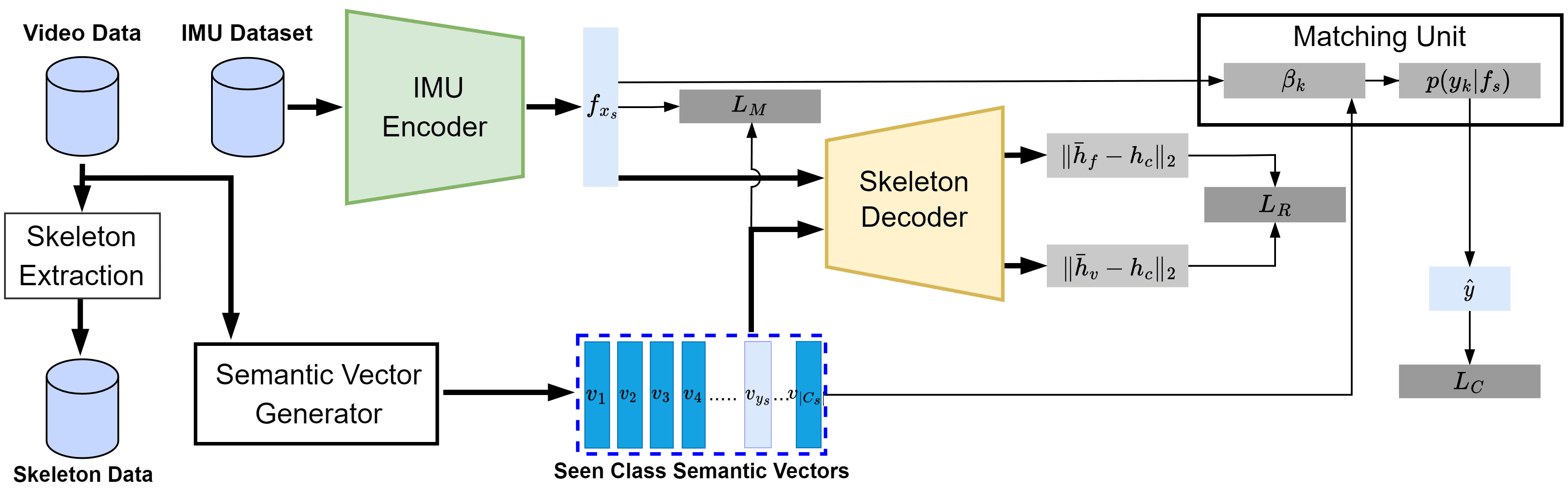}
    \caption{The overview of the training process of SEZ-HARN.} \label{fig:trainin_model_diagram}
\end{figure*}

SEZ-HARN utilizes a Bi-LSTM \cite{p_bilstm}, to generate a vector representation of an IMU sequence. Additionally, it creates class semantic vectors that represent various activity classes by embedding videos of such classes using a pre-trained video encoder. The encoded IMU sequence and class semantic vectors are then fed into the \textit{Matching Unit} to determine the class of the given IMU sequence. Furthermore, the encoded IMU sequence goes through a skeleton decoder to create a skeleton video that explains the predicted class. Fig. ~\ref{fig:inference_model_diagram} shows the inference process of SEZ-HARN. Below we describe the training procedure and each component of SEZ-HARN in detail. Fig. ~\ref{fig:trainin_model_diagram} shows the training process of the proposed model.

Let $C_s$ be the set of seen classes while $C_u$ is the unseen classes set and $C_s \cap C_u = \emptyset $. We denote the training dataset as $D_s$ and the testing dataset as $D_u$. A training sample of SEZ-HARN consists of an IMU sample and its corresponding activity label, $ (x_s, y_s) \in D_s$ where $y_s \in C_s$. $x_s$ is a multivariate time-series:  $ x_s \in \mathcal{R}^{n \times d}$, where $n$ is the sequence length and $d$ is the feature dimension. Given $x_s$, an IMU encoder generates a high-level feature vector $f_{x_s}$. The IMU encoder in SEZ-HARN comprises a Bi-LSTM followed by a dropout, ReLU, and linear layers. Hence, $f_{x_s}$ includes temporal information encoded in the multivariate IMU signal.

SEZ-HARN is designed to learn the relationship between seen and unseen classes by analyzing videos. However, many IMU-based HAR datasets do not have accompanying video data, making it difficult, and sometimes impossible, to collect videos that match the recorded IMU sequences. To overcome this challenge, we utilise public video repositories, such as YouTube, to compile a collection of videos for a specific activity class. Although these videos may not align perfectly with the IMU sequences, they still help SEZ-HARN learn generic patterns of activities as demonstrated in our evaluation study. 

SEZ-HARN uses a pre-trained video encoder to convert videos of human activities into a semantic space. To do this, we feed the encoder a set of $n$ videos related to each activity class $c \in C_s$ and obtain a set of feature vectors. We then find the average of these vectors, which we call the "class semantic vector" of class $c$ or $v_c$.We create a set of class semantic vectors $V = \{v_1, v_2, ..., v_{|C_s|}\}$ and the class semantic vector of $y_s$ is called $v_{y_s}$. Our system, SEZ-HARN, learns the semantic relationship between seen and unseen activity classes by minimizing the L2 distance defined as $L_M$ between $f_{x_s}$ and $v_{y_s}$  as given  (\ref{eq:matching_loss}).  %all symbols defined.

\begin{equation} \label{eq:matching_loss}
    L_M = \|f_{x_s} - v_{y_s} \|_2
\end{equation}

To determine the class of $x_s$, we feed $f_{x_s}$ and $V$ to a \textit{Matching Unit}. It outputs the class of the $v_c$ most similar to $f_{x_s}$ as the class of $x_s$. Matching Unit first projects $f_{x_s}$ onto the unit vector of each class's semantic vector. Let the similarity between $f_{x_s}$ and $v_k \in V$ be $\beta_k$ as denoted by (\ref{eq:simlarity_equation}) where $ k \in {1,2, .., |C_s|}$. %all symbols defined.

\begin{equation} \label{eq:simlarity_equation}
    \beta_k = f_{x_s} \cdot \frac{v_k}{\| v_k \|_2}
\end{equation}

Then Matching Unit applies SoftMax normalization as given in (\ref{eq:softmax_equation}) to derive the probability of class $y_k$ given $x_s$.%all symbols defined.

\begin{equation} \label{eq:softmax_equation}
    P(y_k|x_s) = \frac{exp(\beta_k)}{\sum_{k \in |C_s|}{exp(\beta_k)}}
\end{equation}

The classification objective, $L_C$, is defined using the negative log-likelihood as given in (\ref{eq:cross_loss}).

% \begin{equation} \label{eq:cross_loss}
%     L_C = -\sum_{k \in |C_s|} y_s \log (P(y_k|x_s))
% \end{equation}

\begin{equation} \label{eq:cross_loss}
    L_C = - \log P(y_k = c_s|x_s)
\end{equation}

%all symbols defined.

The SEZ-HARN model extends existing IMU-based ZS-HAR models by incorporating the generation of skeleton videos to explain its decisions. To achieve this, SEZ-HARN utilizes the decoder from the Bidirectional Recurrent Autoencoder-based skeleton autoencoder proposed by Li et al. \cite{p_skeldecoder}. The process begins by selecting a random video from the collected set of videos corresponding to the activity class $c \in C_s$. The video is passed through the BlazePose model \cite{p_blazepose} to extract the coordinates of 25 skeleton key points, including face and finger positions. From these, 12 predominant key points are selected, denoted as $h_c$, which represent the primary skeleton movements associated with the activity class.

SEZ-HARN is trained to reconstruct $h_c$ using the skeleton decoder, guided by the IMU feature vector $f_{x_s}$ and the class semantic vector $v_{y_s}$. The reconstructed skeleton video conditioned on $f_{x_s}$ is denoted as $\bar{h}_f$, while the reconstructed skeleton video conditioned on $v_{y_s}$ is denoted as $\bar{h}_v$. SEZ-HARN is optimized by minimizing the L2 distance between the generated skeleton sequences ($\bar{h}_f$ and $\bar{h}_v$) and the original skeleton movements $h_c$, as shown in (\ref{eq:skel_recon_loss}). This process enables SEZ-HARN to generate skeleton videos corresponding to the predicted class and enhances the mapping between $f_{x_s}$ and $v_{y_s}$.

\begin{equation} \label{eq:skel_recon_loss}
    L_R = \|\bar{h}_f - h_c \|_2 + \|\bar{h}_v - h_c \|_2
\end{equation}

The final objective function of the proposed SEZ-HARN as given in (\ref{eq:comb_loss}) is a linear combination of $L_{M}, L_{C}$ and $L_{R}$.
\begin{equation} \label{eq:comb_loss}
    L = L_{M}+\lambda L_{C} + \alpha L_{R}
\end{equation}
, where $\lambda$ and $\alpha$ are hyper-parameters.
%all symbols defined.

\section{Experimental Study}
\label{sec:experiments}

% This section outlines the details of our experimental study conducted to evaluate the effectiveness of the proposed SEZ-HARN model. Firstly, we provide information about the datasets used in our experiments. Next, we elaborate on the implementation details of the SEZ-HARN model. After that, we present various experiments to evaluate different aspects of the proposed model. First, we compare SEZ-HARN with state-of-the-art ZS-HAR models to assess how well it recognizes unseen classes. Then, we investigate the knowledge transferability of our model from seen to unseen classes and analyze the impact of each loss term in the final objective function. Finally, we evaluate the quality and understandability of the explanations generated to ensure they are realistic and human-understandable.

% \subsection{Benchmark and Auxiliary Datasets}

\subsection{Datasets}

% \subsubsection{IMU Benchmark Datasets}

We use four IMU datasets commonly used for benchmarking ZS-HAR in our experiments. Namely, we use PAMAP2 \cite{d_pamap2}, DaLiAc \cite{d_daliac}, UTD-MHAD \cite{d_utd-mhad} and MHEALTH \cite{d_mhealth}. These datasets contain IMU signals captured by sensors on different body parts, such as the ankle, wrist, and chest. Each sensor provides measurements of acceleration, gyroscope, and magnetometer readings across the X, Y, and Z axes. 
% Typically, these data are collected with a limited number of subjects performing action sequences sequentially. During annotation, various sensor readings are combined and marked with time boundaries for each action.
Table \ref{tab:datasets} shows the summary of the IMU datasets. Overall, these datasets provide a variety of activity recognition challenges, from different numbers of subjects and sensors to various activity types and durations, making them useful for evaluating HAR models.

\begin{table}[!t] 
    \centering
    \small
    \caption{IMU dataset characteristics}
    \setlength\tabcolsep{3pt}
    \begin{tabular}{lccccc}
    \hline
        \textbf{Dataset} & \textbf{Activities} & \textbf{Subjects} & \textbf{Samples} & \textbf{Features} & \textbf{Folds} \\ \hline
        \href{https://archive.ics.uci.edu/dataset/231/pamap2+physical+activity+monitoring}{\textbf{PAMAP2}} & 18 & 9 & 5169 & 54 & 5 \\ 
        \href{https://www.mad.tf.fau.de/research/activitynet/daliac-daily-life-activities/}{\textbf{DaLiAc}} & 13 & 19 & 21844 & 24 & 4 \\ 
        \href{https://personal.utdallas.edu/~kehtar/UTD-MHAD.html}{\textbf{UTD-MHAD}} & 27 & 8 & 861 & 6 & 5 \\ 
        \href{https://archive.ics.uci.edu/dataset/319/mhealth+dataset}{\textbf{MHEALTH}} & 12 & 10 & 2774 & 12 & 4 \\ \hline
    \end{tabular}
    \label{tab:datasets}
\end{table}

%Overall, these datasets provide a variety of activity recognition challenges, from different numbers of subjects and sensors to various activity types and durations, making them useful for evaluating HAR models. Table \ref{tab:datasets} shows the summary of the datasets.

SEZ-HARN builds the semantic space by exploiting videos of activities. However, none of the above datasets, except UTD-MHAD, accompanies video data. Hence, we collected supplementary video datasets from publicly available repositories \cite{d_kinetic400}, such as YouTube, for the PAMAP2, DaLiAc, and MHEALTH datasets. We searched for videos using the activity class label and collected ten videos for each activity. For activities present in multiple datasets (e.g., "walking"), we share the same set of videos across those datasets. To reduce noise and maintain consistency, we ensured each video featured only one subject, with minimal limb cropping and occlusions. We aimed to capture the entire action sequence within a fixed time frame, regardless of the natural speed of the actions. All samples within the same action class performed the same action, with variations only in subject, camera angle, and distance. For example, we selected the action of goalie side jumping for the "Playing Soccer" class in the PAMAP2 dataset. The collected video set for PAMAP2, DaLiAc, and MHEALTH datasets can be found at \url{https://bit.ly/sezharn_videos}.

\subsection{Implementation}

Our experiments use the I3D model \cite{p_i3d} as the video encoder in SEZ-HARN and the decoder of Skeleton Autoencoder proposed in \cite{p_skeldecoder} as the skeleton decoder. The I3D model was pre-trained on the Kinetic-400 dataset \cite{d_kinetic400}, whereas the skeleton autoencoder was pre-trained on the NTURGB 120 dataset \cite{d_nturgb}. To obtain the coordinates of the skeleton key points to fine-tune the Skeleton Autoencoder, we use the BlazePose model \cite{p_blazepose}. 

The activity classes in all four datasets used in our experiments can be categorized into super-classes \cite{p_har_zsl_video}. For example, the 14 activities in the PAMAP2 dataset can be categorized into five super-classes: static, walking, house chores, sports, and sitting, as shown in Table ~\ref{tab:pamap2_supercls}. We employ a $k$-fold evaluation approach to partition the activity classes into seen and unseen sets. The separation strategy, similar to \cite{p_har_zsl_video}, is used for the PAMAP2, DaLiAc, and UTD-MHAD datasets. For the MHEALTH dataset, three unseen classes are randomly chosen for each of the four folds. The activity classes are categorized based on their activity super-class, such as static, dynamic, and sports. Unseen classes are created by randomly selecting activities from each super-class. The k-fold class separation guarantees that each fold's seen and unseen class sets contain at least one sample from each activity super-class. Within each fold, the seen dataset is divided into a 90\% training dataset and a 10\% validation dataset.

SEZ-HARN is implemented using PyTorch \cite{p_torch} and trained on an NVIDIA Tesla T4 GPU or an NVIDIA GeForce RTX 2040 GPU. The ADAM optimizer \cite{p_adam} with a learning rate of $10^{-3}$ is used in training. We train SEZ-HARN for 20 epochs with a batch size of 64. $\lambda$ and $\alpha$ in Equation~\ref{eq:comb_loss} are set to $10^{-2}$ and 0.6, respectively after rigorous hyper-parameter tuning. The hidden size and the LSTM stacks are set to 128 and 2 in the Bi-LSTM-based IMU encoder, while the dropout rate is 0.1. Additional details on the model's implementation and experimentation can be found at \url{https://github.com/SEZ-HARN/SEZ-HARN}.

\begin{table}[!t]
    \caption{Activity super-class definition in the PAMAP2}
    \centering
    %\small
    \begin{tabular}{p{0.2\textwidth}p{0.8\textwidth}}
    %\begin{tabular}{ll}
    \hline
        \textbf{Activity} & \textbf{Action Classes } \\ \hline
        \textbf{Static} & lying, sitting, standing \\ 
        \textbf{Walking} & walking, Nordic walking, ascending stairs, descending stairs \\ 
        \textbf{House chores} & vacuum cleaning, ironing, folding laundry, house cleaning \\ 
        \textbf{Sports} & running, cycling, playing soccer, rope jumping \\ 
        \textbf{Sitting} & watching TV, computer work, car driving \\
    \hline
    \end{tabular}
    \label{tab:pamap2_supercls}
\end{table}

\subsection{Comparative Study}
\label{sec:results}

We compare SEZ-HARN with the state-of-the-art (SOTA) IMU-based ZS-HAR models: MLCLM \cite{p_har_zsl_mcross}, VbZSL \cite{p_har_zsl_video}, and TEZARNet \cite{125PathirageICONIP2023}, in terms of unseen classification accuracy. MLCLM and VbZSL use a Multi-Layer Perceptron on static features extracted from IMU data. MLCLM utilizes word embedding to create the semantic space, whereas VbZSL utilizes video embedding. Like SEZ-HARN, TEZARNet uses a BiLSTM-based architecture and video embedding to make the semantic space, but employs a neighborhood-based unseen class prediction in contrast to SEZ-HARN. However, all these SOTA models are black-box models that cannot explain their decisions. 

Following the current work in ZS-HAR, we use the \textit{average accuracy per class} as the evaluation metric in our experiments. %The average accuracy per class compensates for the class imbalance problem in most of the standard IMU-based HAR datasets. 
Suppose the number of correct predictions for a unseen class $c_u$ is $N^{correct}_{c_u}$ and number of total instances for $c_u$ is $N^{total}_{c_u}$. The average accuracy per class is defined in (\ref{eq:perfM})
%edit here

\begin{equation} \label{eq:perfM}
    \textrm{Average Accuracy per Class} = \frac{1}{|C_u|} \sum_{c_u \in C_u}\frac{N^{correct}_{c_u}}{N^{total}_{c_u}} 
\end{equation}

For MLCLM and TEZARNet models, we refer to the accuracy values reported in the respective papers. Since VbZSL implementation is not publicly available, we use our implementation of VbZSL and train it using the video datasets employed in the SEZ-HARN training process.

The results in Table~\ref{tab:comp} indicate that our model consistently achieves a higher average accuracy per class across all four datasets than MLCLM and VbZSL. Compared to the recent model, TEZARNet, SEZHARN achieves higher or on-par average accuracy per class in all the datasets except PAMAP2. Further, VbZSL lags in performance due to its limited utilisation of temporal information in the IMU data. Furthermore, TEZARNet and SEZ-HARN outperform MLCLM and VbZSL with word embeddings, demonstrating that incorporating video data as auxiliary information in IMU data-based ZS-HAR models improves performance. These results indicate that introducing explainability has not compromised the performance of SEZ-HARN.

\begin{table*}[!t]
\small
    \caption{Comparison of \textit{Average Accuracy per Class} over k-folds for different datasets.}
    \centering
    \begin{tabular}{lcccc}
    \hline
        \textbf{Model} & \textbf{PAMAP2} & \textbf{DaLiAc} & \textbf{UTD-MHAD} & \textbf{MHEALTH} \\ \hline
        MLCLM\cite{p_har_zsl_mcross} & 54.93 & - & - & - \\ 
        VbZSL\cite{p_har_zsl_video}(Video) & 42.20  & 70.60 & 24.84 & 38.80 \\ 
        VbZSL\cite{p_har_zsl_video}(Word) & 47.70 & 60.00 & 32.40 & - \\ 
        TEZARNet\cite{125PathirageICONIP2023} & \textbf{58.27} & 76.10 & \textbf{32.60} & 40.40 \\        \hline
        %SEZ-HARN (Our model) & Video Emb. & 55.20 & \textbf{76.41} & 32.52 & \textbf{46.67} \\ \hline
        SEZ-HARN & 55.20 & \textbf{76.41} & 32.52 & \textbf{46.67} \\ \hline
       % % \hline
       %  Perc. Impr. & & -0.05 & +0.004 & -0.002 & +0.155 \\
 
    \end{tabular}
    \label{tab:comp}
\end{table*}

\subsection{Knowledge Transferability} \label{sec: knw_tranfer_latent}

The success of the ZSL model relies on its ability to transfer knowledge from seen classes to unseen classes, allowing it to recognize new actions based on what it has learned from seen actions. This study evaluates SEZ-HARN's knowledge transferability using IMU feature vectors and skeleton video explanations using the PAMAP2 dataset.

\begin{figure*}[!t]
    \includegraphics[width=\linewidth]{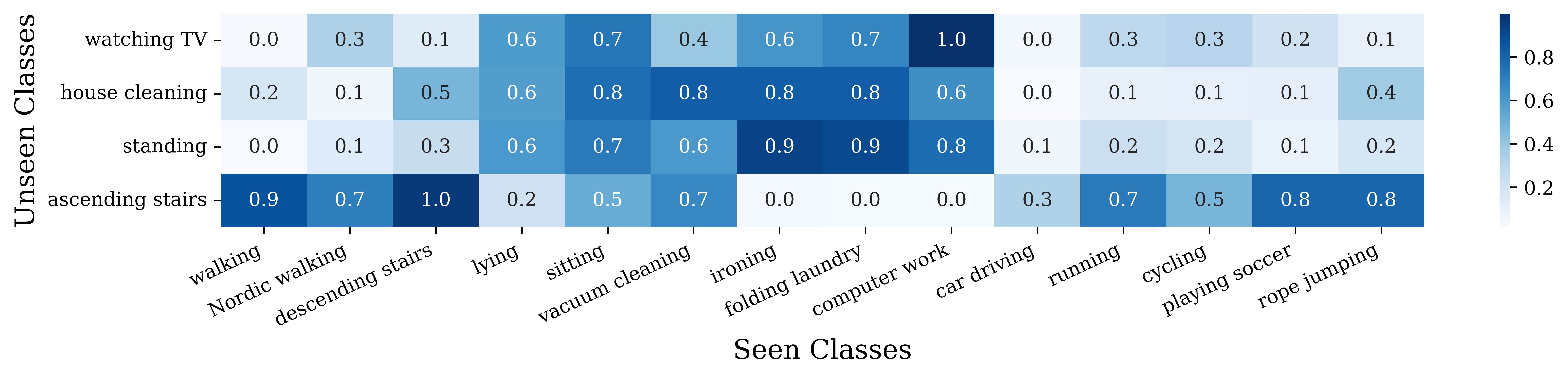}
%    \caption{Seen-Unseen Latent Space Similarity} 
    \caption{The cosine similarity score between seen and unseen classes of the PAMPA2 dataset SEZHARN trained on video embeddings} 
    \label{fig:seen_unseen_heatmap}
\end{figure*}

To assess the knowledge transferability of SEZ-HARN using IMU feature vectors, we extract these vectors for both seen and unseen classes through the trained model. Then, we create a class-IMU-feature vector for each class by calculating the average of each class's IMU feature vectors. Finally, we compute the Cosine Similarity between each pair of seen and unseen class-IMU-feature vectors. The heatmap in Fig.~\ref {fig:seen_unseen_heatmap} shows the similarities between each pair of seen and unseen classes in the PAMAP2 dataset for a single fold.

We observe that similarity scores between seen and unseen classes are significantly higher within the same super-class compared to those across different super-classes, highlighting SEZ-HARN's ability to effectively capture semantic alignment. For instance, "ascending stairs" has cosine similarities of 0.7, 0.9, and 1.0 with "Nordic walking," "walking," and "descending stairs," respectively—all activities within the "walking" super-class. In contrast, "ascending stairs" exhibits much lower cosine similarities of 0.2 and 0.5 with "lying" and "sitting," which belong to the "static" super-class. This clear distinction indicates that SEZ-HARN successfully transfers knowledge from seen to unseen classes by leveraging semantic relationships and maintaining strong super-class alignment.

% In comparison, MLCLM and VbZSL \cite{p_har_zsl_mcross, p_har_zsl_video} show weaker semantic alignment, as seen in Figures ~\ref{fig:mlclm_glove_seen_unseen_heatmap} and 
% ~\ref{fig:vbzsl_seen_unseen_heatmap}. For example, in MLCLM, "ascending stairs" has uniformly high cosine similarities (e.g., 0.9) with both in-super-class activities like "walking" and across-super-class activities like "lying" and "sitting." Similarly, in VbZSL, the distinction between super-classes is less pronounced, with "ascending stairs" showing higher similarities (e.g., 0.7 and 0.8) to "lying" and "sitting" than in SEZ-HARN. In contrast, SEZ-HARN demonstrates a sharper alignment, with markedly high similarities within super-classes (e.g., 1.0 with "descending stairs") and low similarities across super-classes (e.g., 0.2 with "lying"). This distinction underscores SEZ-HARN’s superior ability to encode semantic features that align strongly with action super-classes, thereby enhancing its knowledge transferability and improving its performance on unseen classes.

%WARNING :- ORIGINAL PLACE OF Table 

Next, we evaluate SEZ-HARN's knowledge transferability by analyzing the explanations provided through skeleton movement videos. In ZSL, the explanations should demonstrate how knowledge is transferred from seen to unseen classes ~\cite{geng2021explainable}. Hence, we expect the skeleton video generated by SEZ-HARN explaining the predicted unseen activity to correspond to a seen activity of its super-class. For example, for the "ascending stairs" activity in the PAMAP2 dataset, the generated skeleton movement video should be similar to a known activity in its super-class of "walking." Hence, we evaluate the alignment of the generated skeleton movement videos with the predicted class's super-class. For this evaluation, we introduce a set of new metrics based on Dynamic Time Wrapping (DTW) \cite{p_dtw}.

DTW \cite{p_dtw} is a technique commonly used to measure the similarity between two sequences, such as time series, that may have variations in length or temporal distortions. It is beneficial when comparing sequences with differing speeds or minor temporal shifts or noise are present. The DTW algorithm determines an optimal alignment between the two sequences by warping and stretching their respective time axes. 

In our experiments, we employ DTW with the Mahalanobis distance \cite{p_mahalanobis} to identify the most similar reference sequence for a given sequence. This approach enables an effective comparison and matching of skeleton movements by accounting for both temporal variations and the underlying structural characteristics of the skeletons \cite{p_mahalanobis}.

The DTW algorithm calculates the optimal alignment path and the corresponding similarity score between the two given sequences $X$ of length $n$ and $Y$ of length $m$. The DTW equation is defined as:

\begin{equation} \label{eq:dtw}
    DTW(X, Y) = \min \left( \sum_{i=1}^{n} \sum_{j=1}^{m} d(i, j) \right)
\end{equation}

\begin{equation} \label{eq:mahalanobis}
    d(i, j) = \sqrt{(i - j)^T S^{-1} (i - j)}
\end{equation}

\noindent subject to the following constraints:

\begin{align*}
&DTW(0, 0) = 0 \\
&DTW(i, 0) = \infty \quad for \quad i > 0 \\
&DTW(0, j) = \infty \quad for \quad j > 0 \\
&DTW(i, j) = c(i, j) + \min(DTW(i-1, j),  \\
&\qquad DTW(i, j-1), DTW(i-1, j-1)) \quad for \quad i, j > 0
\end{align*}
% \begin{multline*}
% DTW(i, j) = c(i, j) + \min(DTW(i-1, j), DTW(i, j-1), \\
% DTW(i-1, j-1)) \quad for i, j > 0
% \end{multline*}

\noindent where c(i, j) represents the local cost or dissimilarity measure between elements i and j of sequences X and Y, respectively. The DTW equation computes the minimum cumulative cost path, representing the optimal alignment between the two sequences, under the Covariant matrix $S$ that defines the joints'  relative movement restrictions. By comparing the DTW score with a predefined threshold, we can determine the similarity between the skeleton movement sequences.

\begin{table}[!t]
    \caption{Model explanations based knowledge adaptability experiment results}
    \centering
    %\small
    \begin{tabular}{lcccc}
    \hline
        \textbf{Dataset} & \textbf{TSA} & \textbf{PSA} & \textbf{OA} & \textbf{ADD} \\ \hline
        \textbf{PAMAP2} & 87.8 & 73.3 & 80.3 & 5.77  \\ 
        \textbf{DaLiAc} & 95.8 & 50.2 & 90.4 & 4.34  \\ 
        \textbf{MHEALTH} & 92.3 & 66.6 & 80.9 & 5.93  \\ 
        \textbf{UTD-MHAD} & 57.1 & 31.4 & 44.4 & 8.4 \\ \hline
    \end{tabular}
    \label{tab:inter_tranfer}
\end{table}

Given the generated skeleton movement video of an unseen instance, we calculate its DTW distance to each of the seen class skeleton videos we used for training. The class of the seen skeleton video with the minimum DTW distance is referred to as the "matching seen class." We introduce three metrics: Target Super-class Alignment (TSA), Predicted Super-class Alignment (PSA), and Overall Alignment (OA).

\begin{itemize}
    \item \textbf{Target Super-class Alignment(TSA)}: TSA is calculated when the unseen prediction is correct. It is the percentage of matching seen class belonging to the super-class of the target class of the given unseen instance.
    \item \textbf{Predicted Super-class Alignment (PSA)}: We calculate PSA when the unseen prediction is incorrect. PSA is the percentage of matching seen class belonging to the super-class of the predicted class. This helps us understand how well the explainability aligns with the model's prediction, even when the prediction is incorrect.
    \item \textbf{Overall Alignment (OA)}: We calculate OA without considering the accuracy of the prediction. OA is the percentage of matching seen class belonging to the super-class of the predicted class irrespective of the correctness of the model prediction.
\end{itemize}

The TSA, PSA, and OA values for all datasets are shown in Table~\ref{tab:inter_tranfer}. We also show the average DTW distance (ADD) between the generated explanation skeleton movement video and the matching seen class skeleton video. The results show that the generated explanations align well with the predicted class's super-class skeleton videos. This indicates that SEZ-HARN has successfully learned the semantic relationship between seen and unseen classes.

\begin{figure*}[!ht]
    \centering
        \includegraphics[width=0.85\linewidth]{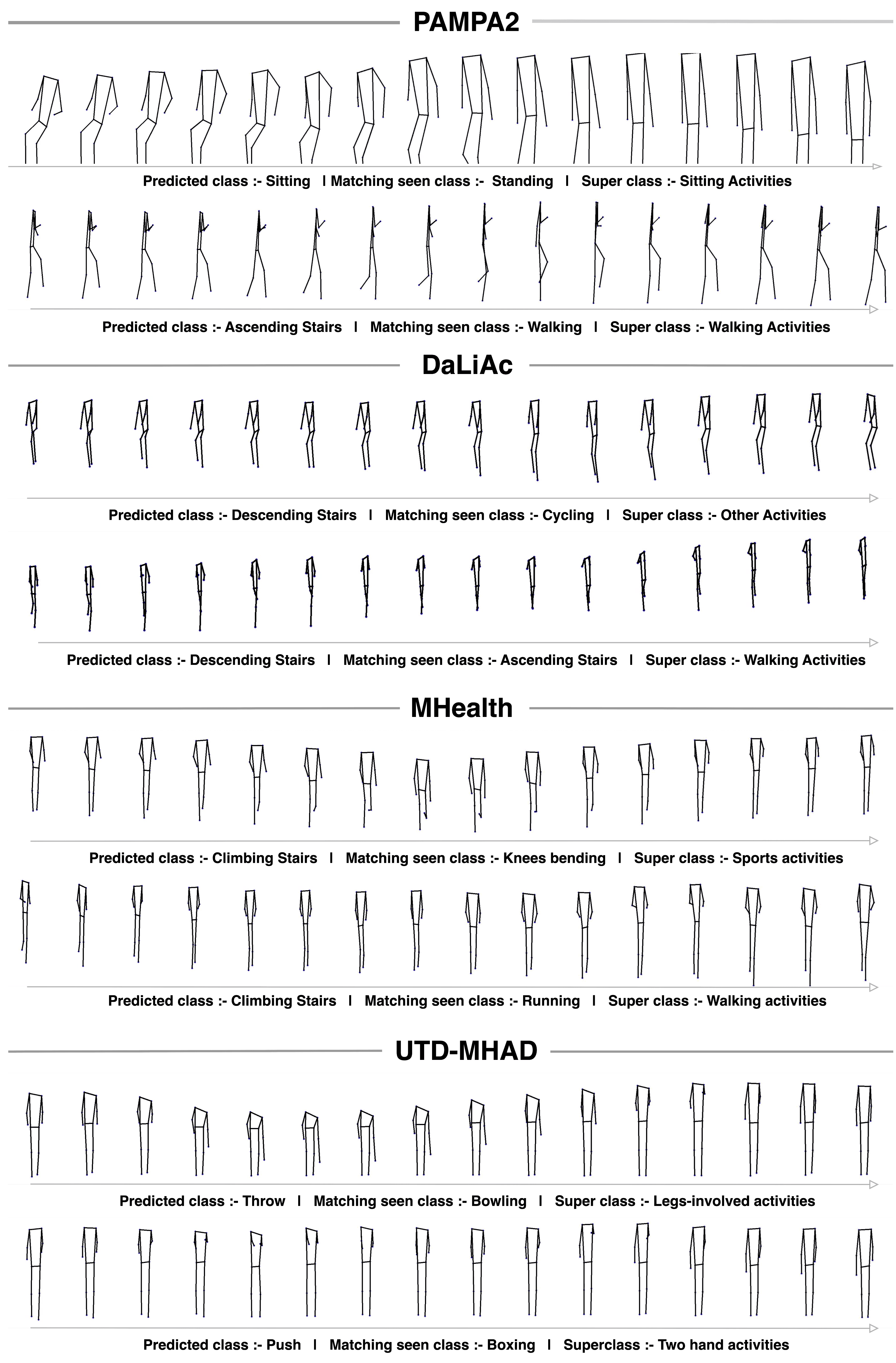}
        \caption{Generated skeleton movement Video Samples for four datasets} 
        \label{fig:skeleton_images}
\end{figure*}

Figure ~\ref{fig:skeleton_images} shows sample skeleton sequences generated by SEZ-HARN explaining the unseen predictions for all four datasets. %For each skeleton sequence, we show the predicted class, the matching seen class, and the super-class of the predicted class.
We observe that the generated skeletons closely resemble the corresponding skeletons from the matching seen classes. Further, the generated skeletons accurately capture the structure of the human skeleton at each frame, displaying smooth joint movements and having minimal ghosting or shaking artefacts. However, the generated skeleton movement videos for UTD-MHAD show relatively lower similarity to reference videos, consistent with results in Table~\ref{tab:inter_tranfer}. This discrepancy can be attributed to the low number of samples in the dataset and the limited number of principal body movements in the skeleton videos used to train the model for this dataset. 

\subsection{Realism of SEZ-HARN Explanations}

To be effective and useful, SEZ-HARN's explanatory skeleton movement videos should display smooth joint movements and minimal ghosting or shaking artifacts; they should be \textit{"realistic"}. We assess the videos' "realism" by evaluating whether they follow principles of body movement and exhibit relative joint movements similar to the original target skeleton action. The raw skeleton movement matrix is used for evaluation, as it contains the skeleton joint coordinates generated by the model for each time frame of the action sequence. These coordinates are then utilized to create the corresponding skeleton action video.

We use Discrete Frechet Distance (DFD) \cite{p_dfd}, to evaluate the realism of the generated skeleton movements compared to the original skeleton movements of the matching classes. DFD is a valuable metric for assessing the similarity between curves or trajectory data. It measures the minimum movement needed for one sequence to traverse another, considering the relative positions and distances between the points in the sequences. Suppose $P$ and $Q$ represent the two sequences being compared. Then the DFD is defined as,

\begin{equation} \label{eq:dfd}
    DFD(P, Q) = \min \left( \max_{\pi} \left( \min \left( \max \left( \left\| p_i - q_{\pi(i)} \right\| \right) \right) \right) \right)
\end{equation}

\noindent where \(p_i\) and \(q_i\) represent the points in the $P$ and $Q$ sequences, respectively, and \(\pi\) represents a permutation of indices that determines the matching between the points. The DFD is computed by finding the optimal matching \(\pi\) that minimizes the maximum distance between corresponding points in the sequences. The employed DFD-based method calculates a dissimilarity between the generated skeleton sequence and the matching seen class skeleton sequence. The DFD ranges from 0 to infinity, where smaller values indicate a closer resemblance to the natural movement of the matching seen class's skeleton sequence.

Our study uses DFD to interpret skeleton movement as a set of joint movement curves. The distinctive characteristics of the DFD make it suitable for analyzing the extent to which a set of skeleton joint movements should be adjusted to align with a reference set of joint movement curves, taking into account both spatial and temporal aspects of the data.

Table \ref{tab:inter_natural} shows the mean of DFD for explanatory skeleton videos generated for all four datasets. The results show that SEZ-HARN produces highly realistic skeleton movement videos for the PAMAP2, DaliAc, and MHEALTH. The relatively higher score in UTD-MHAD can be attributed to the skeleton decoder generating novel skeleton movements due to the low sample count relative to the class count and a significantly lower percentage of reference skeleton data exhibiting principal body movements.

\begin{table}[!t]
    \caption{Model explanation realism evaluation}
    \centering
    %\small
    %\begin{tabular}{p{1.53cm}p{1.3cm}p{1.15cm}p{0.75cm}p{1.8 cm}}
     \begin{tabular}{lcccc}
    %\begin{tabular}{lcccc}
    \hline
        \textbf{Metrics} & \textbf{MHEALTH} & \textbf{PAMAP2} & \textbf{DaLiAc} & \textbf{UTD-MHAD} \\ \hline
        \textbf{DFD-Mean} & 0.507 & 0.445 & 0.359 & 7.971  \\ 
        \textbf{DFD-std} & 0.031 & 0.033 & 0.145 & 9.965 \\ \hline
    \end{tabular}
    \label{tab:inter_natural}
\end{table}

\subsection{Human Understandability of SEZ-HARN Explanations}

We conduct a user study to assess the human understandability of the generated skeleton movement videos. First, we randomly select thirteen unseen IMU samples from the PAMAP2 dataset covering all five super-classes. The selected IMU sample set contains at least one sample from each super-class. Then we feed these IMU samples to SEZ-HARN and collect the explanation skeleton movement videos generated by SEZ-HARN. In the user study, participants are asked to identify the super-class corresponding to each skeleton movement video and provide a confidence value for their selection, ranging from 0 to 5, where 5 indicates the highest confidence. This evaluation aims to measure the clarity of the explanations provided by the participants' super-class identification accuracy. The user study can be found at \url{https://forms.gle/Fyw7xY3rikzE7UvC7}.

% \begin{figure}[!t]
% \centering
% \begin{minipage}[t]{0.37\linewidth}
%     \centering
%     \includegraphics[width=\linewidth]{dfd_interpret_figureV5.jpg}
%     \caption{Joint movement trajectory deviation analysis between reference (black) and generated (pink) skeleton sequence.} 
%     \label{fig:dfd_vis}
% \end{minipage}
% \hfill
% \begin{minipage}[t]{0.59\linewidth}
%     \centering
%     \includegraphics[width=\linewidth]{survey_resultV8.jpg}
%     \caption{Hinton plot illustrating survey participants' choices, with color indicating selection percentage and size reflecting confidence level.} 
%     \label{fig:survey_vis}
% \end{minipage}
% \end{figure}

% \begin{figure}[!t]
% \centering
%     \includegraphics[width=0.8\linewidth]{dfd_interpret_figureV5.jpg}
%     \caption{Joint movement trajectory deviation analysis between reference (black) and generated (pink) skeleton sequence.} 
%     \label{fig:dfd_vis}
% \end{figure}

\begin{figure}[!ht]
\centering
    \includegraphics[width=0.8\linewidth]{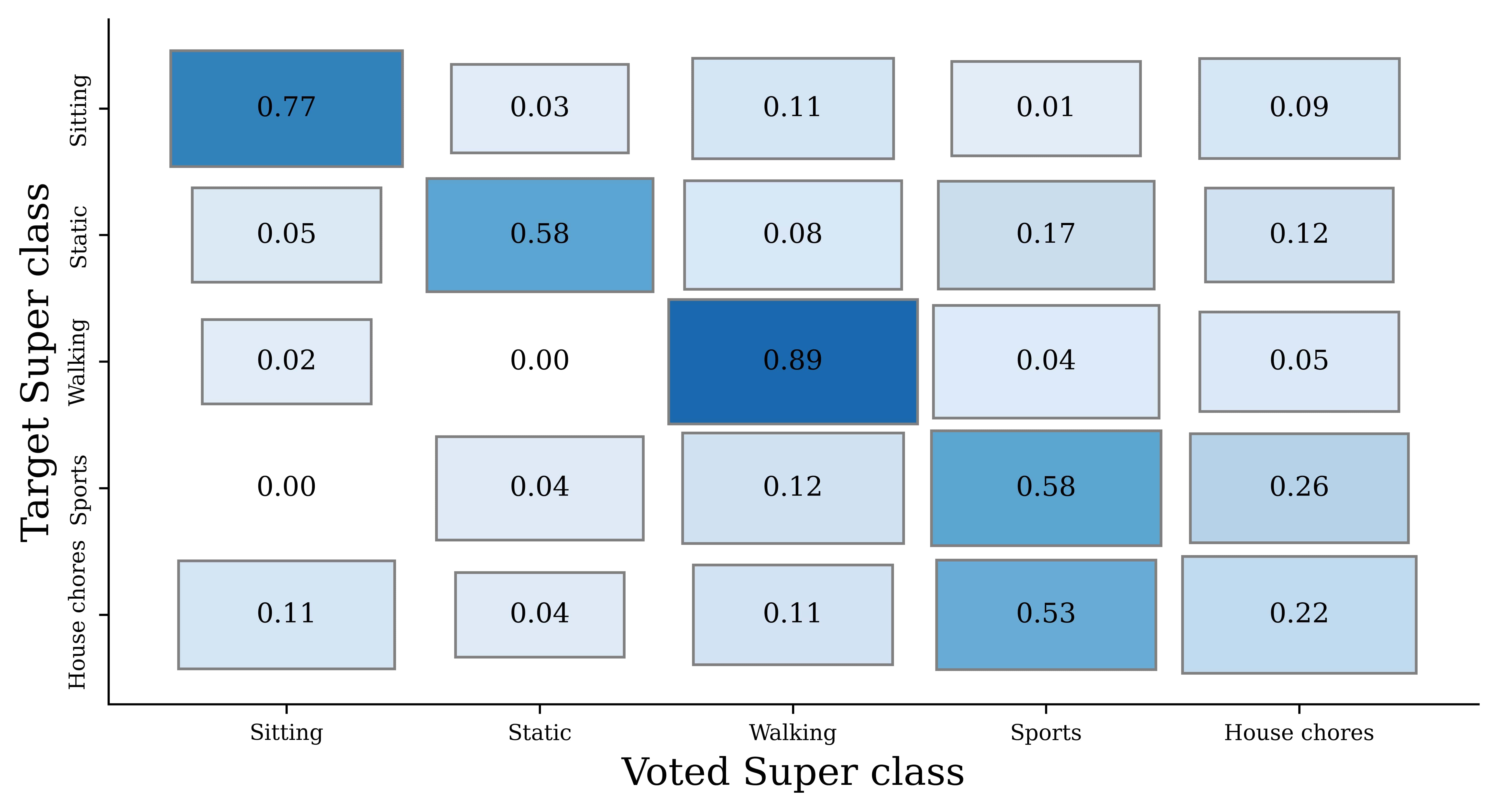}
    \caption{Hinton plot illustrating survey participants' choices, with color indicating selection percentage and size reflecting confidence level} 
    \label{fig:survey_vis}
\end{figure}

% We collected responses from 53 volunteers with diverse levels of ML knowledge, including beginners, novices, intermediate users, and researchers. All respondents were between 18 and 40 years old, with most being undergraduates.

Fifty-three volunteers with diverse levels of ML knowledge, from beginners, to researchers, participated in our survey. All participants were between 18 and 40 years old, with the majority being undergraduates.

Fig. \ref{fig:survey_vis} shows the participants' response accuracy heatmap for selecting super-classes corresponding to the provided skeleton movement videos. The results indicate significant accuracy in identifying the correct super-class for all categories, except for the "house chores" super-class. Despite the majority of incorrect responses for the "house chores" super-class, the average confidence scores for the correct responses remain consistently high across all five super-classes, averaging around 4. This demonstrates that the generated videos are clear and comprehensible, allowing participants to understand the representative action with high confidence. Moreover, based on empirical evidence, we attribute the lower accuracy for the "house chores" super-class to the inherent complexity and diversity of actions typically associated with this category.

\section{Discussion}
\label{sec:discussion}

We propose SEZ-HARN—Self-Explainable Zero-shot Human Activity Recognition Network—extending recent ZS-HAR models such as VbZSL \cite{p_har_zsl_video} and TEZARNet \cite{125PathirageICONIP2023}. SEZ-HARN constructs the semantic space using auxiliary activity videos, leveraging their rich motion information. A key innovation is its ability to generate self-explanatory skeleton movement videos, addressing the explainability limitations in existing ZS-HAR models \cite{p_har_zsl_wemb, p_har_zsl_mcross, p_har_zsl_video}. While prior supervised sensor-based HAR models \cite{p_xai_har,p_dexar,dubey2022xai} have incorporated post-hoc explanation methods such as SHAP \cite{p_shap} and Grad-CAM \cite{selvaraju2017grad}, these approaches are not specifically designed for zero-shot models and often produce explanations that are difficult for non-expert users to interpret. SEZ-HARN bridges this gap by generating visually intuitive skeleton movement videos, enabling transparent and user-friendly explanations.

We evaluated SEZ-HARN on four publicly available IMU-based HAR datasets. As shown in Table~\ref{tab:comp}, it consistently outperforms MLCLM and VbZSL, and achieves comparable accuracy to TEZARNet—except on PAMAP2, where performance is slightly lower. These results indicate that self-explainability in SEZ-HARN does not compromise recognition performance. Moreover, constructing the semantic space from video data and leveraging temporal features from IMU signals enhances recognition of unseen activities.

To assess explanation quality, we analyzed realism and interpretability of the generated skeleton videos. Realism, measured via Discrete Fréchet Distance, was high across datasets, with the exception of UTD-MHAD, likely due to its smaller sample size and limited motion diversity. A user study confirmed that the videos were perceived as intuitive and interpretable, supporting SEZ-HARN’s potential for real-world deployment.

Nevertheless, SEZ-HARN inherits dataset limitations such as limited activity diversity, class imbalance, and actor bias, which may affect generalizability. Future work should explore more diverse and augmented auxiliary data to strengthen zero-shot performance. Other future directions include improving the explanation mechanism to highlight salient motion patterns that influence model decisions, developing interactive and multi-modal explanations, and enhancing scalability to accommodate broader and more complex HAR domains.

\section{Conclusion}
\label{sec:conclusion}

We present a Zero-Shot Human Activity Recognition (ZS-HAR) model that addresses two key challenges for adapting IMU-based HAR models to real-world scenarios: the limited availability of labeled data and the lack of transparency in existing models. The proposed approach leverages video data to learn semantic relationships between seen and unseen classes while generating skeleton movement videos to explain its decisions. Extensive experiments on four benchmark datasets—PAMAP2, DiLiAc, UTD-MHAD, and MHEALTH—demonstrate that the model effectively captures the semantic alignment between seen and unseen classes, outperforming state-of-the-art ZS-HAR models in all datasets except PAMAP2. Furthermore, the generated explanations are both realistic and intuitive, ensuring they are easily understandable to human users.

\section*{Declarations}

\begin{itemize}
    \item Availability of data and materials:
The IMU-HAR datasets used in this study are publicly available and can be accessed from their respective official repositories, as cited in the manuscript. Additionally, we have collected videos for each IMU-HAR dataset, which can be found at \url{https://bit.ly/sezharn_videos}.

 \item Competing interests:
The authors declare no conflict of interest.

 \item Funding:
This research did not receive any specific grant from funding agencies in the public, commercial, or not-for-profit sectors.

\item Ethics approval: 
Not applicable

\item  Consent to participate:
 Not applicable

\item  Code availability: The source code is publicly available at \url{https://github.com/SEZ-HARN/SEZ-HARN}

\item Author’s Contribution:
\textbf{Devin Y. De Silva:} Methodology, Software, Validation, Investigation, Data Curation, Writing - Original Draft, Visualization
\textbf{Sandareka Wickramanayake:} Conceptualization, Methodology, Writing - Review \& Editing, Supervision, Project administration
\textbf{Dulani Meedeniya:} Writing - Review \& Editing, Supervision
\textbf{Sanka Rasnayaka:} Writing - Review \& Editing, Supervision

All authors have read and agreed to the published version of the manuscript.

\end{itemize} 
\noindent

\section*{Declaration of generative AI and AI-assisted technologies in the writing process}

During the preparation of this work, the authors used Grammarly and ChatGPT in order to improve the grammar, clarity, and flow of the manuscript. After using this tool/service, the author(s) reviewed and edited the content as needed and take full responsibility for the content of the publication.
\bibliographystyle{elsarticle-num-names} 
\bibliography{main}

\appendix






%%\bibitem[Lamport(1994)]{lamport94}
 %% Leslie Lamport,
%%  \textit{\LaTeX: a document preparation system},
%%  Addison Wesley, Massachusetts,
%%  2nd edition,
%%  1994.

%%\end{thebibliography}
\end{document}